\documentclass{llncs}

\usepackage{multicol}
\usepackage{bbm}
\usepackage{url}
\usepackage{amssymb,array}
\usepackage{enumitem}

\usepackage{amsmath}

\usepackage{algorithm}
\usepackage{algpseudocode}

\usepackage{placeins}
\usepackage{graphicx}
\graphicspath{{Images/}}
\usepackage{caption}
\usepackage{subcaption}
\usepackage{tikz} 

\usepackage[colorlinks=true,linkcolor=blue,urlcolor=black, bookmarksdepth=2]{hyperref}
\usepackage{bookmark}

\usepackage{makeidx}  
\usepackage[draft]{todonotes}

\begin{document}
%

%
\pagestyle{headings}  

\mainmatter              

\title{Mosquito Detection with Neural Networks: The Buzz of Deep Learning}
%

%
\author{Ivan Kiskin\inst{1,2}  \and Bernardo P\'erez Orozco\inst{1,2} \and Theo Windebank\inst{1,3} \and Davide Zilli\inst{1,2} \and Marianne Sinka\inst{4,5}, Kathy Willis\inst{4,5,6}  \and Stephen Roberts\inst{1,2}}
%
%
%
\institute{University of Oxford, Department of Engineering, Oxford OX1 3PJ, UK, \\
\and
\email{\textrm{\{}ikiskin, ber, dzilli, sjrob\textrm{\}}@robots.ox.ac.uk}, \and \email{theo.windebank@stcatz.ox.ac.uk},  \\
\and University of Oxford, Department of Zoology, Oxford OX2 6GG, UK, \\
\and \email{\{marianne.sinka, kathy.willis\}@zoo.ox.ac.uk} \\
\and Royal Botanic Gardens, Kew, Richmond, Surrey, TW9 3AE, UK.
}

\maketitle              
\begin{abstract}

Many real-world time-series analysis problems are characterised by scarce data. Solutions typically rely on hand-crafted features extracted from the time or frequency domain allied with classification or regression engines which condition on this (often low-dimensional) feature vector. The huge advances enjoyed by many application domains in recent years have been fuelled by the use of deep learning architectures trained on large data sets. This paper presents an application of deep learning for acoustic event detection in a challenging, data-scarce, real-world problem. Our candidate challenge is to accurately detect the presence of a mosquito from its acoustic signature. We develop convolutional neural networks (CNNs) operating on wavelet transformations of audio recordings. Furthermore, we interrogate the network's predictive power by visualising statistics of network-excitatory samples. These visualisations offer a deep insight into the relative informativeness of components in the detection problem. We include comparisons with conventional classifiers, conditioned on both hand-tuned and generic features, to stress the strength of automatic deep feature learning. Detection is achieved with performance metrics significantly surpassing those of existing algorithmic methods, as well as marginally exceeding those attained by individual human experts. The data and software related to this paper are available at \url{http://humbug.ac.uk/kiskin2017/}.

\end{abstract}

\section{Introduction}
Mosquitoes are responsible for hundreds of thousands of deaths every year due to their capacity to vector lethal parasites and viruses, which cause diseases such as malaria, lymphatic filariasis, zika, dengue and yellow fever \cite{WHO2016,WHO2014factsheet}. Their ability to transmit diseases has been widely known for over a hundred years, and several practices have been put in place to mitigate their impact on human life. Examples of these include insecticide-treated mosquito nets \cite{lengeler2004insecticide,bhatt2015} and sterile insect techniques \cite{alphey2010sterile}. However, further progress in the battle against mosquito-vectored disease requires a more accurate identification of  species and their precise location -- not all mosquitoes are vectors of disease, and some non-vectors are morphologically identical to highly effective vector species.
%
Current surveys rely either on human-landing catches or on less effective light traps. In part this is due to the lack of cheap, yet accurate, surveillance sensors that can aid mosquito detection.
Our work uses the acoustic signature of mosquito flight as the trigger for detection. Acoustic monitoring of mosquitoes proves compelling, as the insects produce a sound both as a by-product of their flight and as a means for communication and mating. Detecting and recognising this sound is an effective method to locate the presence of mosquitoes and even offers the potential to categorise by species. Nonetheless, automated mosquito detection presents a fundamental signal processing challenge, namely the detection of a weak signal embedded in noise. Current detection mechanisms rely heavily on domain knowledge, such as the likely fundamental frequency and harmonics, and extensive hand-crafting of features -- often similar to traditional speech representation. With impressive performance gains achieved by a paradigm shift to deep learning in many application fields, including bioacoustics \cite{joly2016lifeclef}, an opportunity emerges to leverage these advances to tackle this problem.

Deep learning approaches, however, tend to be effective only once a critical number of training samples has been reached \cite{chen2014flying}. Consequently, data-scarce problems are not well suited to this paradigm. As with many other domains, the task of data labelling is expensive in both time requirement for hand labelling and associated ambiguity -- namely that multiple human experts will not be perfectly concordant in their labels. Furthermore, recordings of free-flying mosquitoes in realistic environments are scarce \cite{Mukundarajan2017} and hardly ever labelled.

This paper presents a novel approach for classifying mosquito presence using scarce training data. Our approach is based on a convolutional neural network classifier conditioned on wavelet representations of the raw data. The network architecture and associated hyperparameters are strongly influenced by constraints in dataset size. To assess our performance, we compare our methods with well-established classifiers, as well as with simple artificial neural networks, trained on both hand-crafted features and the short-time Fourier transform. We show that our classifications are made more accurately and confidently, resulting in a precision-recall curve area of 0.909, compared to 0.831 and 0.875 for the highest scoring traditional classifier and dense-layer neural network respectively. This performance is achieved on a classification task where only 70\,\% of labels are in full agreement amongst four domain experts. We achieve results matching, and even surpassing, human expert level accuracy. The performance of our approach allows realistic field deployments to be made as a smartphone app or on bespoke embedded systems.

This paper is structured as follows. Section \ref{sec:Context} addresses related work, explaining the motivation and benefits of our approach. Section \ref{sec:method} details the method we adopt. Section \ref{sec:ExperDetails} describes the experimental setup, in particular emphasising data-driven architectural design decisions. Section \ref{sec:results} highlights the value of the method. We visualise and interpret the predictions made by our algorithm on unseen data in Section \ref{subsec:visual} to help reveal informative features learned from the representations and verify the method.  Finally, we suggest further work and conclude in Section \ref{sec:Conclusion}.

\section{Related Work}
\label{sec:Context}
%
The use of artificial neural networks in acoustic detection and classification of species dates back to at least the beginning of the century, with the first approaches addressing the identification of bat echolocation calls \cite{parsons2000}. Both manual and algorithmic techniques have subsequently been used to identify insects \cite{chesmore2004automated,zilli2014hidden}, elephants \cite{clemins2002automatic}, delphinids \cite{oswald2003}, and other animals. The benefits of leveraging the sound animals produce -- both actively as communication mechanisms and passively as a results of their movement -- is clear: animals themselves use sound to identify prey, predators, and mates. Sound can therefore be used to locate individuals for biodiversity monitoring, pest control, identification of endangered species and more.

This section will therefore review the use of machine learning approaches in bioacoustics, in particular with respect to insect recognition. We describe the traditional feature and classification approaches to acoustic signal detection. In contrast, we also present the benefit of feature extraction methods inherent to current deep learning approaches. Finally, we narrow our focus down to the often overlooked wavelet transform, which offers significant performance gains in our pipeline.


\subsection{Insect Detection}
\label{subsec:InsectDetect}
Real-time mosquito detection provides a method to combat the transmission of lethal diseases, mainly malaria, yellow fever and dengue fever. Unlike \textit{Orthoptera} (crickets and grasshoppers) and \textit{Hempitera} (e.g. cicadas), which produce strong locating and mating calls, mosquitoes (\textit{Diptera, Culicidae}) are much quieter. The noise they emit is produced by their wingbeat, and is affected by a range of different variables, mainly species, gender, age, temperature and humidity. In the wild, wingbeat sounds are often overwhelmed by ambient noise. For these reasons, laboratory recordings of mosquitoes are regularly taken on tethered mosquitoes in quiet or even soundproof chambers, and therefore do not represent realistic conditions. 

Even in this data-scarce scenario, the employment of artificial neural networks has been proven successful for a number of years. In \cite{chesmore2004automated} a neural network classifier was used to discriminate four species of grasshopper recorded in northern England, with accuracy surpassing 70\,\%. Other classification methods include Gaussian mixture models \cite{Potamitis2007,Pinhas2008} and hidden Markov models \cite{leqing2010insect,zilli2014hidden}, applied to a variety of different features extracted from recordings of singing insects.

Chen et al. \cite{chen2014flying} attribute the stagnation of automated insect detection accuracy to the mere use of acoustic devices, which are allegedly not capable of producing a signal sufficiently clean to be classified correctly. As a consequence, they replace microphones with pseudo-acoustic optical sensors, recording mosquito wingbeat through a laser beam hitting a phototransistor array -- a practice already proposed by Moore et al. \cite{moore1986automated}. This technique however relies on the ability to lure a mosquito through the laser beam. 

Independently of the technique used to record a mosquito wingbeat frequency, the need arises to be able to identify the insect's flight in a noisy recording. The following section reviews recent achievements in the wider context of acoustic signal classification. 




\subsection{Feature Representation and Learning}

\label{sub:dl}
The process of automatically detecting an acoustic signal in noise typically consists of an initial preprocessing stage, which involves cleaning and denoising the signal itself, followed by a feature extraction process, in which the signal is transformed into a format suitable for a classifier, followed by the final classification stage. 
Historically, audio feature extraction in signal processing employed domain knowledge and intricate understanding of digital signal theory \cite{humphrey2013feature}, leading to hand-crafted feature representations. 

Many of these representations often recur in the literature. A powerful, though often overlooked, technique is the wavelet transform, which has the ability to represent multiple time-frequency resolutions \cite[Ch. 9]{akay1998time}. An instantiation with a fixed time-frequency resolution thereof is the Fourier transform. The Fourier transform can be temporally windowed with a smoothing window function to create a Short-time Fourier transform (STFT). Mel-frequency cepstral coefficients (MFCCs) create lower-dimensional representations by taking the STFT, applying a non-linear transform (the logarithm), pooling, and a final affine transform. A further example is presented by Linear Prediction Cepstral Coefficients (LPCCs), which pre-emphasise low-frequency resolution, and thereafter undergo linear predictive and cepstral analysis \cite{ai2012classification}.  

Detection methods have fed generic STFT representations to standard classifiers \cite{potamitis2014classifying}, but more frequently complex features and feature combinations are used, applying dimensionality reduction to combat the curse of dimensionality \cite{lee2009unsupervised}. Complex features (e.g. MFCCs and LPCCs) were originally developed for specific applications, such as speech recognition, but have since been used in several audio domains \cite{li2001classification}.


On the contrary, the deep learning approach usually consists of applying a simple, general transform to the input data, and allowing the network to both learn features and perform classification. This enables the models to learn salient, hierarchical features from raw data. The automated deep learning approach has recently featured prominently in the machine learning literature, showing impressive results in a variety of application domains, such as computer vision \cite{krizhevsky2012imagenet} and speech recognition \cite{lee2009unsupervised}. However, deep learning models such as convolutional and recurrent neural networks are known to have a large number of parameters and hence typically require large data and hardware resources. 
Despite their success, these techniques have only recently received more attention in time-series signal processing.

A prominent example of this shift in methodology is the BirdCLEF bird recognition challenge. The challenge consists of the classification of bird songs and calls into up to 1500 bird species from tens of thousands of crowd-sourced recordings. The introduction of deep learning has brought drastic improvements in mean average precision (MAP) scores. The best MAP score of 2014 was 0.45 \cite{goeau2015lifeclef}, which was improved to 0.69 the following year when deep learning was introduced, outperforming the closest scoring hand-crafted method that scored 0.58 \cite{joly2016lifeclef}. The impressive performance gain came from the utilisation of well-established convolutional neural network practice from image recognition. By transforming the signals into STFT spectrogram format, the input is represented by 2D matrices, which are used as training data. Alongside this example, the most widely used base method to transform the input signals is the STFT  \cite{sainath2015deep,gwardys2014deep,potamitis2016deep}.

However, to the best of our knowledge, the more flexible wavelet transform is hardly ever used as the representation domain for a convolutional neural network. As a result, in the following section we present our methodology, which leverages the benefits of the wavelet transform demonstrated in the signal processing literature, as well as the ability to form hierarchical feature representations for deep learning.



%
\section{Method}
\label{sec:method}
We present a novel wavelet-transform-based convolutional neural network architecture for the detection of mosquitoes' flying tone in a noisy audio recording. We explain the wavelet transform in the context of the algorithm, thereafter describing the neural network configurations and a range of traditional classifiers against which we assess performance. The key steps of the feature extraction and classification pipeline are given in Algorithm \ref{alg:Detection}.  

\subsection{The Wavelet Transform}
\label{subsec:wavSTFT}
%
%
\begin{algorithm}[]
	
	\caption{Detection Pipeline}
	\label{alg:Detection}
	\begin{algorithmic}[1]
	\State Load $N$ labelled microphone recordings $x_1(t), x_2(t), \ldots, x_N(t)$.
    \State Take{} transform with $h_1$ features such that we form a feature tensor $\mathbf{X_\text{train}}$ and corresponding label vector $\mathbf{y}_\text{train}$:
       
    $$\mathbf{X}_\textrm{train} \in \mathbb{R}^{N_S
 			\times h_1 \times w_1}, \mathbf{y}_\textrm{train} \in \mathbb{R}^{N_S \times 2},$$
    where $N_s$ is the number of training samples formed by splitting the transformed recordings into 2D `images' with dimensions $h_1 \times w_1$. 
    \label{algline:transform}

	\label{line:method:stft}
  	\State Train classifier on $\mathbf{X}_\textrm{train}, \mathbf{y}_\textrm{train}$.
	\State For test data, $\mathbf{X}_\textrm{test}$, neural network outputs a prediction $y_{i,\textrm{pred}}$ for each class $C_i$: \{$C_0 = \textrm{non-mosquito}$, $C_1 = \textrm{mosquito}$\}, where
			 $$ 0 \leq y_{i,\textrm{pred}}(\mathbf{x}) \leq 1, \quad \textrm{such that} \quad \sum_{i=1}^{n} y_{i,\textrm{pred}}(\mathbf{x}) = 1. $$
	\end{algorithmic}
\end{algorithm}

As an initial step, we extract the training data into a format suitable for the classifier. We choose to use the continuous wavelet transform (CWT) due to its successful application in time-frequency analysis \cite{daubechies2011synchrosqueezed} (Step \ref{algline:transform} of Algorithm \ref{alg:Detection}). Given the direct relationship between the wavelet scale and centre frequency, we use the bump wavelet \cite{vialatte2009bump}, expressed in the Fourier domain as:
\begin{equation} 
\Psi(s\omega) = \exp \left(1 - \frac{1}{1 - (s\omega - \mu)^2/\sigma^2}\right) 
   \mathbbm{I}[(\mu - \sigma)/s, (\mu + \sigma)/s],
\label{eq:bump}   
\end{equation}
where $\mathbbm{I}[\cdot]$ is the indicator function and $s$ is the wavelet scale. High values of $\mu$, as well as small values of $\sigma$, result in a wavelet with superior frequency localisation but poorer time localisation. 

\subsection{Neural Network Configurations}
%


A convolutional layer $H_{\text{conv}}:\mathbb{R}^{h_1\times w_1 \times c} \rightarrow \mathbb{R}^{h_2\times w_2 \times N_k}$ with input tensor $\mathbf{X} \in\mathbb{R}^{h_1\times w_1 \times c}$ and output tensor $\mathbf{Y} \in \mathbb{R}^{h_2\times w_2 \times N_k}$ is given by the sequential application of $N_k$ learnable convolutional kernels $\mathbf{W}_{p} \in \mathbb{R}^{k\times k}, p < N_k$ to the input tensor. Given our single-channel $(c=1$) input representation of the signal $\mathbf{X} \in\mathbb{R}^{h_1\times w_1 \times 1}$ and a single kernel $\mathbf{W}_{p}$, their 2D convolution $\mathbf{Y}_k$ is given by \cite[Ch. 9]{Goodfellow-et-al-2016}:
\begin{equation} 
    \mathbf{Y}_k(i, j) = \mathbf{X}\ast\mathbf{W}_p = \sum_{i'}\sum_{j'}\mathbf{X}(i-i',j-j')\mathbf{W}_p(i',j').
\end{equation}
The $N_k$ individual outputs are then passed through a non-linear function $\phi$ and stacked as a tensor $\mathbf{Y}$. Conventional choices for the activation $\phi$ include the sigmoid function, the hyperbolic tangent and the rectified linear unit (ReLU).

A fully connected layer $H_{\text{FC}}:\mathbb{R}^m \rightarrow \mathbb{R}^n$ with input $\mathbf{x}\in\mathbb{R}^m$ and output $\mathbf{y}\in\mathbb{R}^n$ is given by $\mathbf{y} = H_{\text{FC}}(\mathbf{x}) = \phi(\mathbf{Wx} + \mathbf{b})$, where $\{\mathbf{W}, \mathbf{b}\}$ are the learnable parameters of the network and $\phi$ is the activation function of layer, often chosen to be non-linear.

\begin{figure}[t]
\centering
 \begin{subfigure}[]{.2\textwidth}
    \includegraphics[page = 1, width=1.0\linewidth]{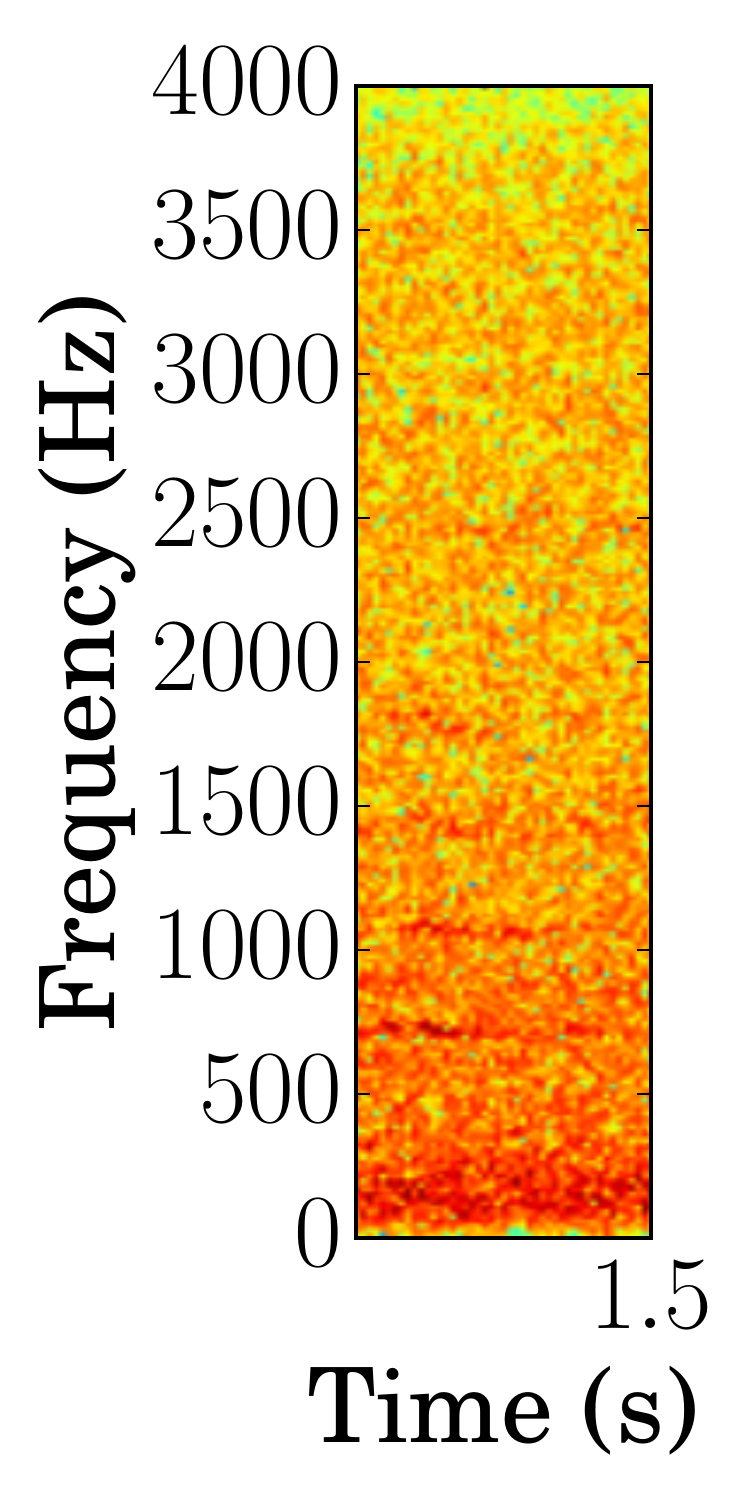}
    \label{fig:subfig:convspec}
  \end{subfigure}%
  \begin{subfigure}[]{.8\textwidth}
    \includegraphics[page = 1, width=1.0\linewidth]{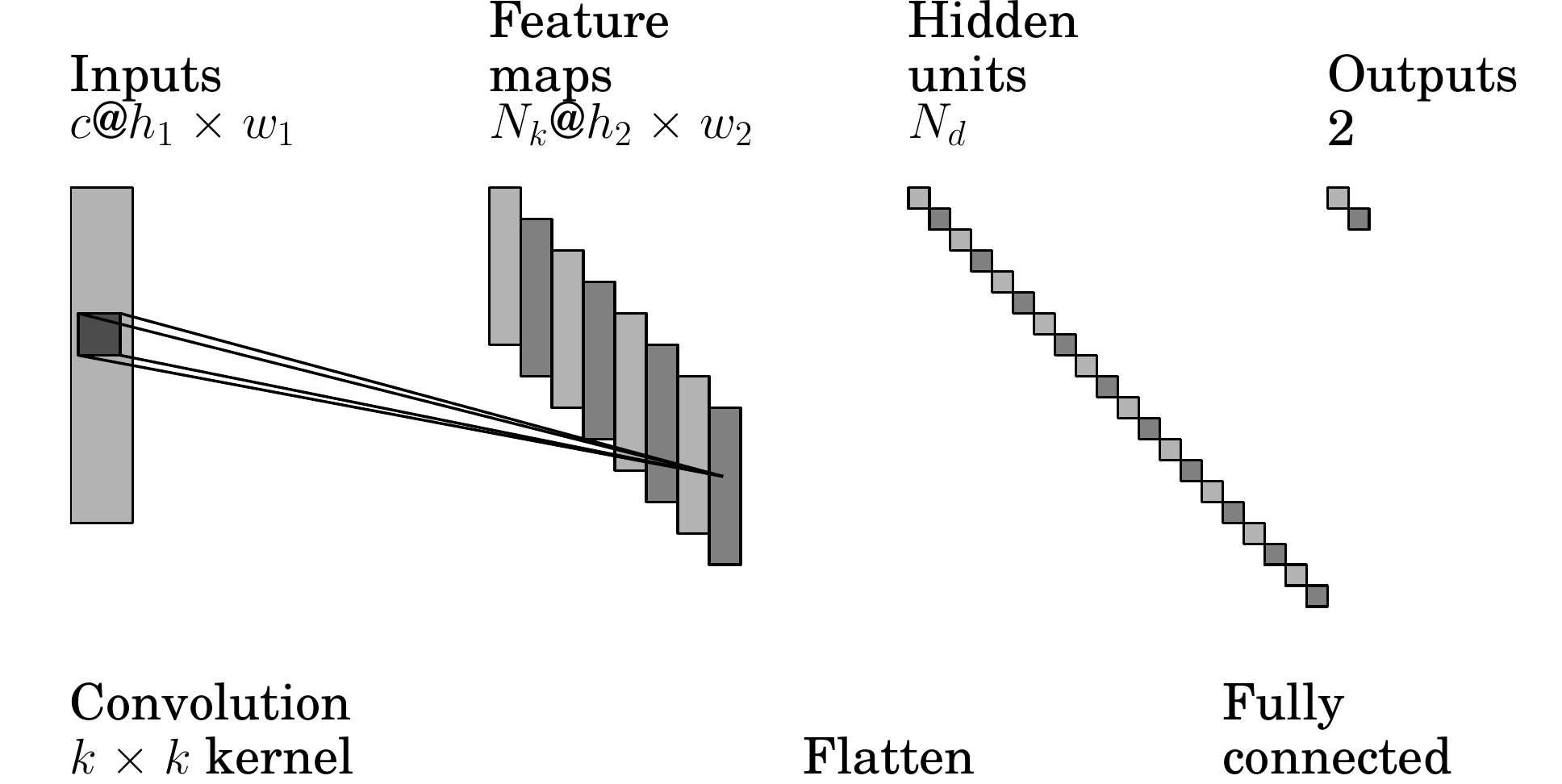}
  \label{fig:subfig:convdiagram}
   \end{subfigure}

\caption{The CNN pipeline. 1.5 s wavelet spectrogram of mosquito recording is partitioned into images with $c=1$ channels, of dimensions $h_1 \times w_1$. This serves as input to a convolutional network with $N_k$  filters with kernel $\mathbf{W}_p \in \mathbb{R}^{k\times k}$.  Feature maps are formed with dimensions reduced to $h_2 \times w_2$ following convolution. These maps are fully connected to $N_d$ units in the dense layer, fully connected to 2 units in the output layer.}
\label{fig:CNN}
\end{figure}

The data size constraint results in an architecture choice (Figure \ref{fig:CNN}) of few layers and free parameters. To prevent overfitting, our network comprises an input layer connected sequentially to a single convolutional layer and a fully connected layer, which is connected to the two output classes with dropout \cite{srivastava2014dropout} with $p=0.5$. Rectified Linear Units (ReLU) activations are employed based on their desirable training convergence properties \cite{krizhevsky2012imagenet}. Finally, potential candidate hyperparameters are cross-validated to determine an appropriate model, as detailed in Section \ref{subsec:xvalparams}.

\begin{figure}
\begin{center}
\begin{tikzpicture}
[   cnode/.style={draw=black,fill=#1,minimum width=3mm,circle},
]

    \node[cnode=black,label=0:$y_2$] (sK) at (7,-4) {};
    \node[cnode=black,label=0:$y_1$] (s1) at (7,-2) {};
    \node[label=0:$\mathrm{outputs}$] at (7,-3) {};
    \node at (0,-3) {$\vdots$};
    \node at (2.5,-3) {$\vdots$};
    \node at (5,-3) {$\vdots$};

    \foreach \x in {1,...,3}
    {   \pgfmathparse{\x<3 ? \x : "D"} 
        \node[cnode=blue,label=180:$x_{\pgfmathresult}$] (x-\x) at (0,{-\x/0.75)}) {};
        \pgfmathparse{\x<3 ? \x : "L"} 
        \node[cnode=gray,label=90:$z^{(1)}_{\pgfmathresult}$] (z-\x) at (2.5,{-\x/0.75)}) {};

        \pgfmathparse{\x<3 ? \x : "M"} 
        \node[cnode=gray,label=90:$z^{(2)}_{\pgfmathresult}$] (p-\x) at (5,{-\x/0.75)}) {};
        \draw (p-\x) -- node[above,sloped,pos=0.3] {${\scriptstyle w^{(3)}_{1\pgfmathresult}}$} (s1);

    } 

    \foreach \x in {1,...,3}
    {   \foreach \y in {1}
        {   
    \pgfmathparse{\x<3 ? \x : "D"}        
        \draw (x-\x) --  node[above,sloped,pos=0.3] {$ {\scriptstyle w^{(1)}_{1\pgfmathresult}}$} (z-\y);
        }
    }

    \foreach \x in {1,...,3}
    {   \foreach \y in {1}
        {   
    \pgfmathparse{\x<3 ? \x : "L"}        
        \draw (z-\x) --  node[above,sloped,pos=0.3] {$ {\scriptstyle w^{(2)}_{1\pgfmathresult}}$} (p-\y);
        }
    }
\end{tikzpicture}
\end{center}
\caption{MLP architecture. For clarity the diagram displays connections for a few units. Each layer is fully connected with ReLU activations. Input dimensions $D = h_1 \times w_1 $. Number of hidden units in the first and second layers labelled $L$ and $M$ respectively.}
\label{fig:NN}

\end{figure}
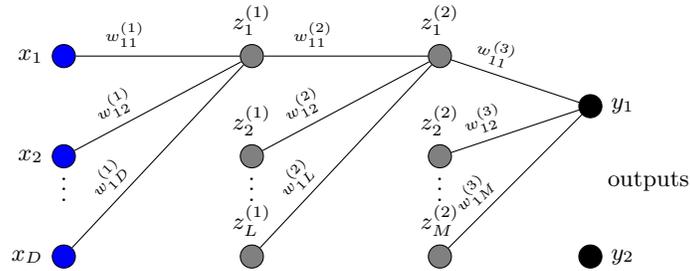

Using conventional multi-layer perceptrons (MLPs) one may simply collapse the matrix $\mathbf{X}$ into a single column vector $\mathbf{x}$. Unlike their convolutional counterparts, MLPs are not explicitly asked to seek relationships among adjacent neurons. Whereas this may provide the model with more flexibility to find relationships between seemingly distant nodes, convolutional layers formally make the model acknowledge that units are correlated in space. Without this assumption, MLPs will look for sets of weights in a space in which this constraint has not been made explicit. Our MLP architecture, chosen for comparison with the CNN, is illustrated in Figure \ref{fig:NN}. The network omits the convolutional layer, taking the form of an input layer followed by two fully connected layers, with dropout with $p=0.5$ on the connections to the output nodes.


\subsection{Traditional Classifier Baseline}
As a baseline, we compare the neural network models with more traditional classifiers that require explicit feature design. We choose three candidate classifiers widely used in machine learning with audio: random forests (RFs), naive Bayes' (NBs), and support vector machines using a radial basis function kernel (RBF-SVMs). Their popularity stems from ease of implementation, reasonably quick training, and competitive performance \cite{silva2013applying}, especially in data-scarce problems. 

We have selected ten features: mel-frequency cepstrum slices, STFT spectrogram slices,  mel-frequency cepstrum coefficients, entropy, energy entropy, spectral entropy, flux, roll-off, spread, centroid, and the zero crossing rate (for a detailed explanation of these features, see for example the open-source audio signal analysis toolkit by \cite{giannakopoulos2015pyaudioanalysis}). To select features optimally, we have applied both recursive feature elimination (RFE) and principal component analysis (PCA), and also cross-validated each feature individually. By reducing redundant descriptors we can improve classification performance in terms of both speed and predictive ability, confirmed by the cross-validation results in Section \ref{subsec:xvalparams}.

\section{Experimental Details}
\label{sec:ExperDetails}
\subsection{Data Annotation}
\begin{figure}[t]
\centering
\includegraphics[width=1\textwidth]{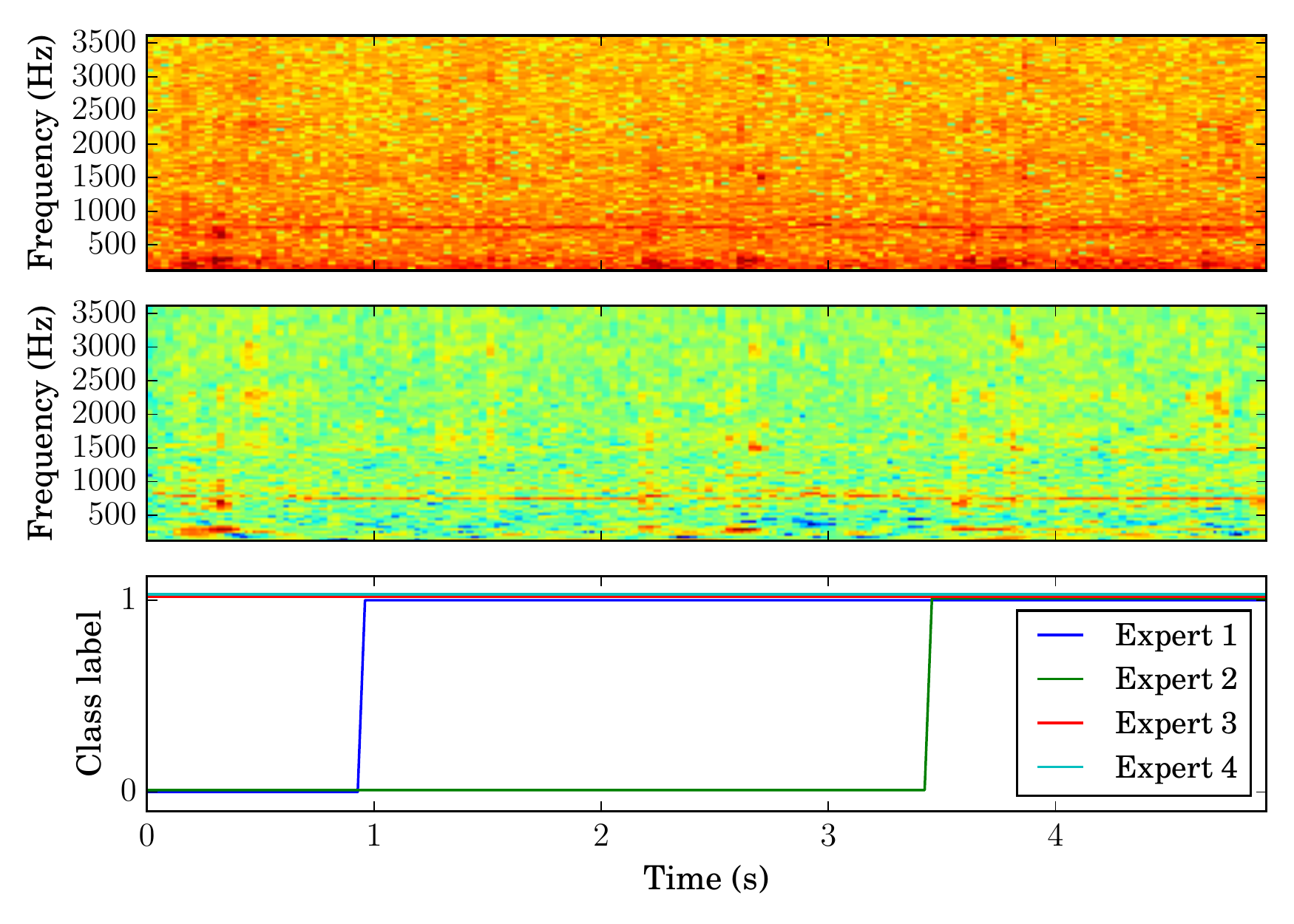}
\caption{STFT (top) and wavelet (middle) representations of signal with $h_1 = 256$ frequency bins and wavelet scales respectively. Corresponding varying class labels (bottom) as supplied by human experts. The wavelet representation shows greater contrast in horizontal constant frequency bands that correspond to the mosquito tone.}
\label{fig:DataLabel}
\end{figure}

The data used here were recorded in January 2016 within culture cages containing both male and female \emph{Culex quinquefasciatus} \cite{bhattacharya2016southern}. The females were not blood-fed and both sexes were maintained on a diet of 10\,\%\,w/v sucrose solution. Figure \ref{fig:DataLabel} shows a frequency domain excerpt of a particularly faint recording in the windowed frequency domains. 
For comparison we also illustrate the wavelet scalogram taken with the same number of scales as frequency bins, $h_1$, in the STFT. We plot the logarithm of the absolute value of the derived coefficients against the spectral frequency of each feature representation.

The signal is sampled at $F_s = 8 $\,kHz, which limits the highest theoretically resolvable frequency to $4$\,kHz due to the Nyquist limit. Figure \ref{fig:DataLabel} (lower) shows the classifications within $y_i = \{0,1\}$: absence, presence of mosquito, as labelled by four individual human experts. Of these, one particularly accurate label set is taken as a gold-standard reference to both train the algorithms and benchmark with the remaining experts. The resulting label rate is given as $F_l = 10$\,Hz. The labels are up-sampled to match the spectral feature frequency, $F_\textrm{spec}$, which is calculated as $F_\textrm{spec} = F_s /h_1 $, provided the overlap between windowed Fourier transforms in samples is half the number of Fourier coefficients.



\subsection{Parameter Cross-Validation}
\label{subsec:xvalparams}
In this section we report the design and parameter considerations that we used cross-validation to estimate. The available 57 recordings were split into 37 training and 20 test signals, creating approximately 6,000 to 60,000 training samples, for window widths $w_1 = 10$ and $w_1 = 1$ samples, respectively. Both neural networks were trained with a batch size of 256 for 20 epochs, according to validation accuracy in conjunction with early stopping criteria.

We start with the CNN and note that the characteristic length scale of the signal determines the choice of slice width. For musical extracts, or bird songs, it is crucial to capture temporal structure. This favours taking longer sections, allowing an appropriate convolutional receptive field in the time domain (along the $x$-axis). A mosquito tone is relatively consistent in frequency over time, so shorter slices are likely to provide a larger training set without loss of information per section. We thus restrict ourselves to dividing the training data into 320\,ms fixed width samples ($w_1 = 10$). When choosing the filter widths to trial, we note that spectrogram samples are correlated in local regions and will contain harmonics that are non-local. The locality is confined to narrow frequency bands, as well as through time (along the $y$ and $x$-axes respectively). Taking this into account, we arrive at the cross-validation grid and results of Table \ref{tab:xval}.

For the MLP, we choose to cross-validate the narrowest training sample width $w_1 = 1$, and the CNN architecture sample width $w_1 = 10$ forming a column vector $\mathbf{x}_\mathrm{train} \in \mathbb{R}^{h_1  w_1 \times 1}$ for each training sample.
We then estimate the optimal number of hidden units as given in Table \ref{tab:xval}.

The traditional classifiers are cross-validated with PCA and RFE dimension reduction as given by $n, m$ in Table \ref{tab:xval}. The best performing feature set for all traditional classifiers is the set extracted by cross-validated recursive feature elimination as in \cite{guyon2002gene}, outperforming all PCA reductions for every classifier-feature pair. The result is a feature set that we denote as $\text{RFE}_{88}$ which retains 88 dimensions from the ten original features which spanned 304 dimensions ($F_{10}  \in \mathbb{R}^{304} $).

\begin{table}[t]
\centering
\caption{Cross-validation results. Optimal hyperparameters given in bold.}
\label{tab:xval}
\begin{tabular}{lll}
\hline\noalign{\smallskip}
Classifier & Features  & Cross-validation grid         \\ 
\noalign{\smallskip}
\hline
\noalign{\smallskip}
CNN          & STFT   & $k \in \{2,\mathbf{3},4,5\}, N_k \in \{8,16,\mathbf{32}\}, N_d \in \{16,64,\mathbf{128},256\}$ \\
CNN          & Wavelet & $k \in \{2,3,4,\mathbf{5}\}, N_k \in \{8,16,\mathbf{32}\}, N_d \in \{16,64,\mathbf{128},256\}$\\
MLP          & STFT   & $ w_1 \in \{1,\mathbf{10}\}, L \in \{8, 256, 1028, \mathbf{2056}\}, M \in \{\mathbf{64}, 512, 1024\} $ \\
MLP          & Wavelet & $w_1 \in \{1,\mathbf{10}\}, L \in \{8, \mathbf{256}, 1028, 2056\}, M \in \{64, 512, \mathbf{1024}\} $ \\ 
\hline
\noalign{\smallskip}
NB, RF, SVM          & $F_{10}  \in \mathbb{R}^{304} $  & $ \text{PCA} \in \mathbb{R}^{N}, N \in 0.8^n \times 304, n \in \{0,1,\ldots,12\},$ \\
&& $\text{RFE} \in \mathbb{R}^{M}, M \in 304 - 8m, m \in \{0, 1, \ldots, \mathbf{27}, \ldots  35\}.  $\\

\hline
\end{tabular}
\end{table}


\section{Classification Performance}
\label{sec:results}

The performance metrics are defined at the resolution of the extracted features and presented in Table \ref{tab:results}. We emphasise that the ultimate goal is deployment in fieldwork on smartphones or embedded devices. The device will be in constant \emph{listening} mode, and mainly consume power during the data \emph{write} mode that is initiated by signal detections. A high true negative rate (TNR) is very desirable for this application, as preventing false positive detections leads to critical conservation of battery power. Taking this into account, we highlight four key results. 

\begin{table}[t]
\centering
\caption{Summary classification metrics. The metrics are evaluated from a single run on test data, following 10-fold cross-validation of features and hyperparameters on training dataset.}
\label{tab:results}
\begin{tabular}{lllllll}
\hline\noalign{\smallskip}
Classifier & Features & $F_1$ score & TPR & TNR & ROC area & PR area \\ 
\noalign{\smallskip}
\hline\hline
\noalign{\smallskip}
MLP        & STFT     & 0.751  		& 0.65                & 0.96   	&  0.858		  & 0.830  \\
MLP        & Wavelet  &  0.745       &    0.63        &    0.97  &    0.921		  & 0.875     \\
CNN        & STFT     &   0.779   &    0.69     		    & 0.96	&  0.871	      &   0.853    \\
CNN        & Wavelet  &   \textbf{0.817}      &      0.73       & \textbf{0.97} &    \textbf{0.952}    &       \textbf{0.909}  \\
Naive Bayes      & STFT  &   0.521    &    0.65       & 0.74 &    0.743  &     0.600  \\
Naive Bayes      & RFE\textsubscript{88}  &   0.484     &     0.51        & 0.83 &   0.732  &     0.414   \\
Random Forest        & STFT &   0.674    &      0.69    & 0.89 &   0.896   &       0.733  \\
Random Forest        & RFE\textsubscript{88}  &   0.710      &    0.68      & 0.93 &    0.920   &       0.800   \\
SVM       & STFT  &   0.685      &      \textbf{0.83}     & 0.81 &   0.902   &      0.775     \\
SVM       & RFE\textsubscript{88}     &      0.745    & 0.73 & 0.93 &    0.928   &       0.831    \\
\hline\hline
CNN, median filter     & Wavelet  &   0.854      &      0.78     & 0.98 &    0.970   &       0.939    \\
\hline
Expert 1       & N/A  &   0.819      &      0.89    & 0.85 &    0.873   &       0.843    \\
Expert 2       & N/A  &   0.856      &      0.92     & 0.88 &    0.901   &       0.873    \\
Expert 3       & N/A  &   0.852      &      0.77     & 0.98 &    0.874   &       0.901    \\
\hline
\end{tabular}
\end{table}

\begin{figure}[bh!]
  \begin{subfigure}[t]{.5\textwidth}
    \centering
    \includegraphics[width=1.\linewidth]{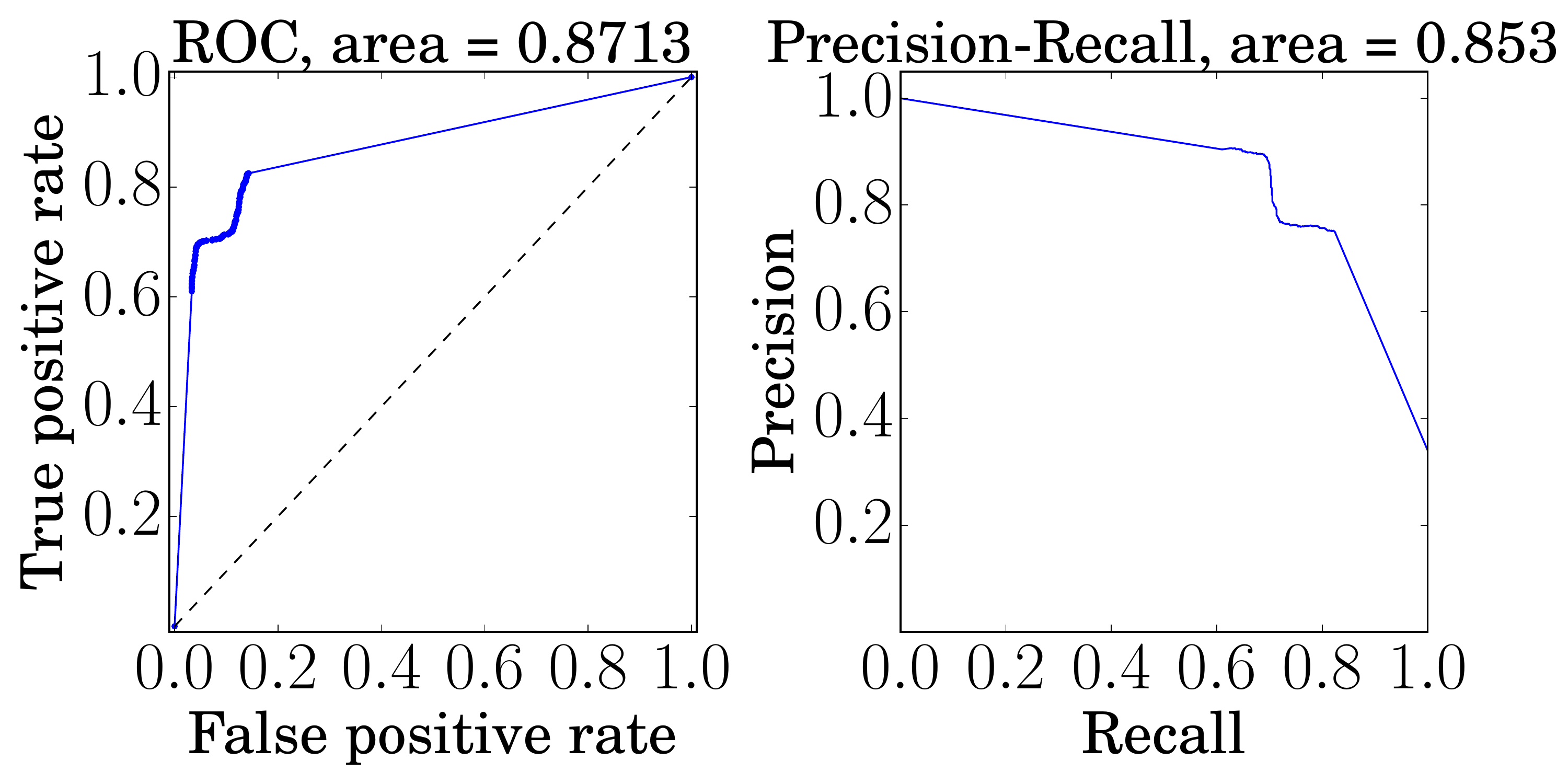}
  \end{subfigure}%
  \begin{subfigure}[t]{.5\textwidth}
  \centering
    \includegraphics[width=1.\linewidth]{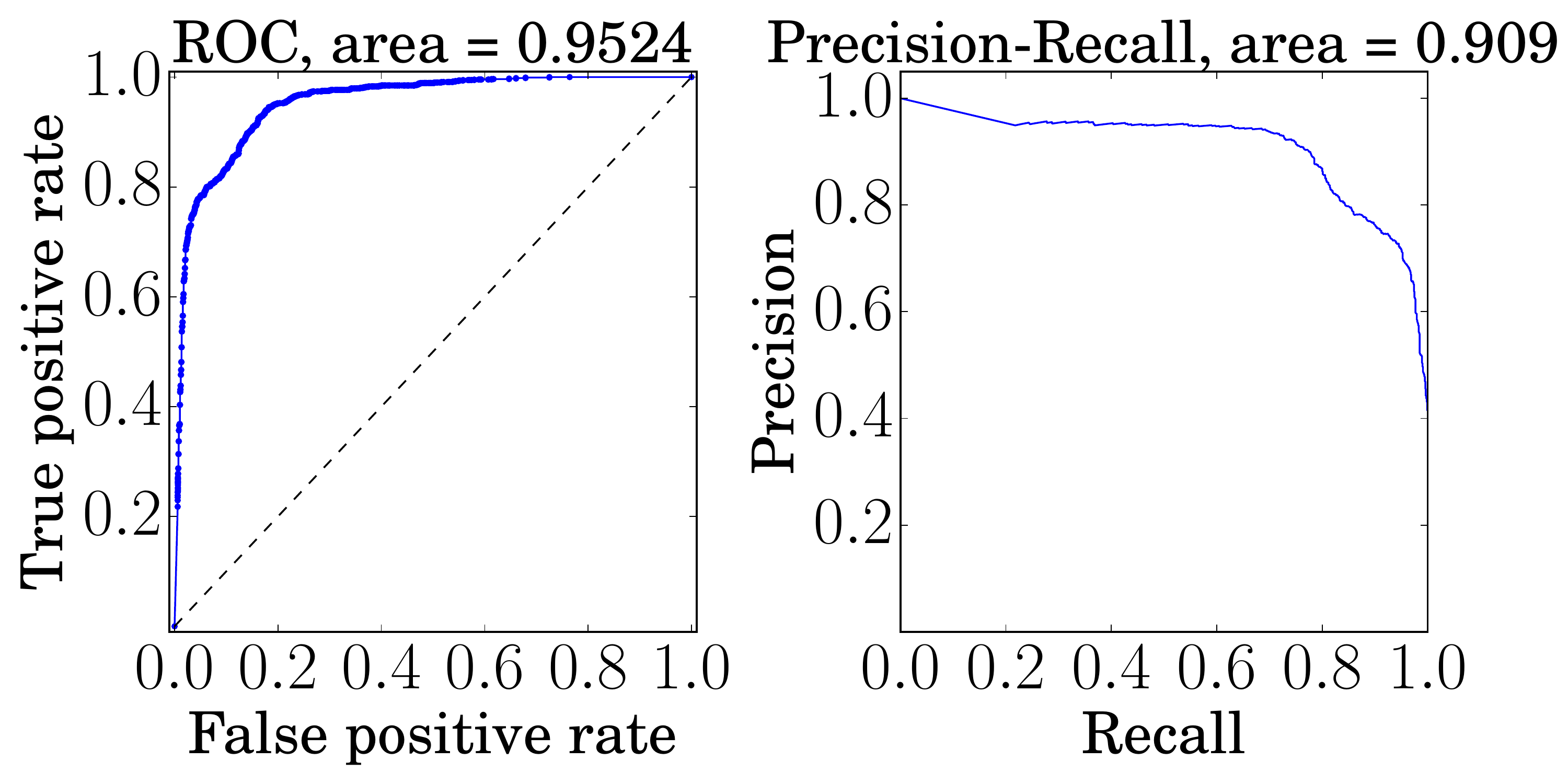}
  \end{subfigure}
  \begin{subfigure}[t]{.5\textwidth}
    \centering
    \includegraphics[width=1.0\linewidth]{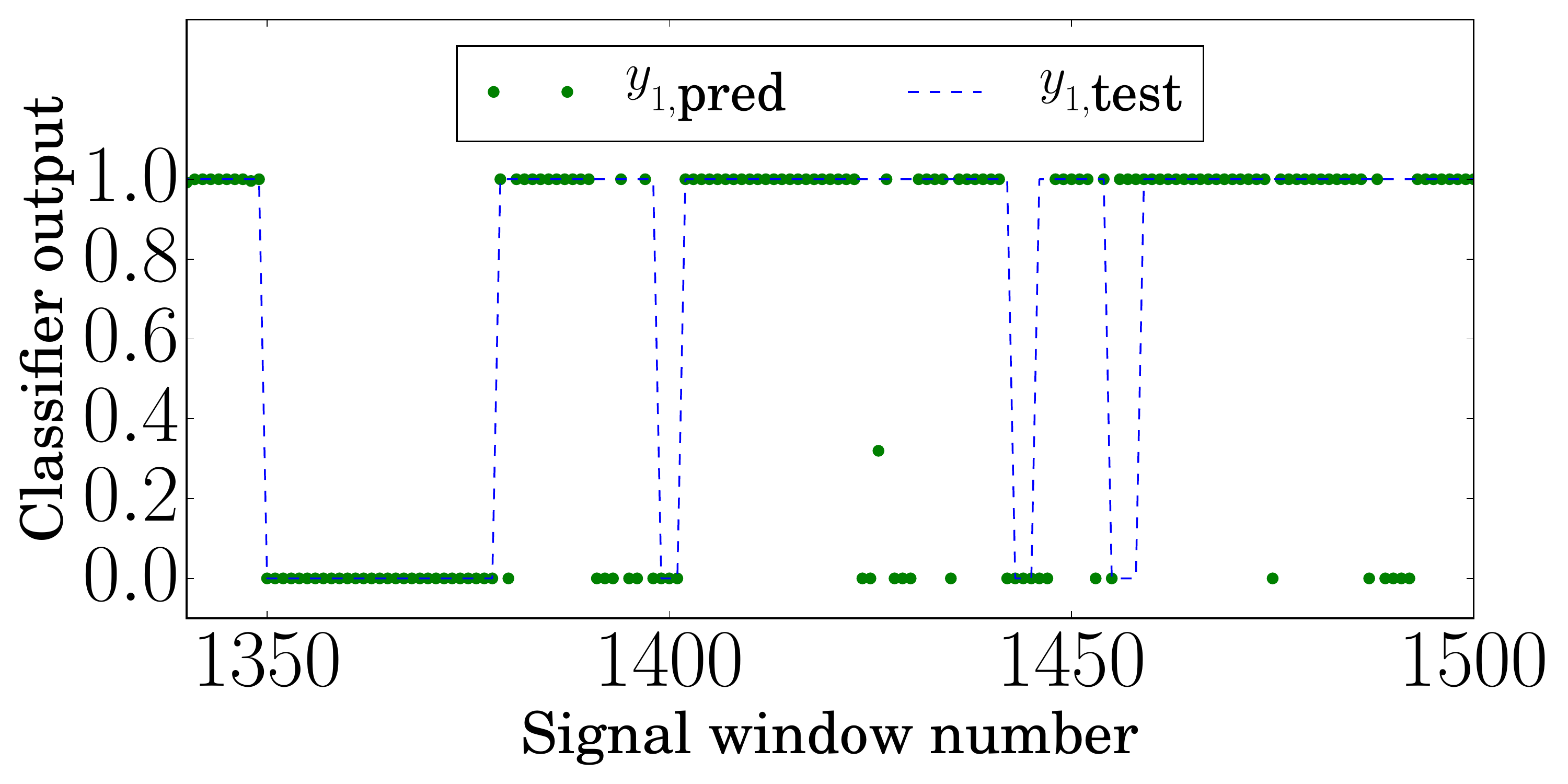}
    \caption{Spectrogram}
    \label{fig:subfig:a}
  \end{subfigure}%
  \begin{subfigure}[t]{.5\textwidth}
  \centering
    \includegraphics[width=1.\linewidth]{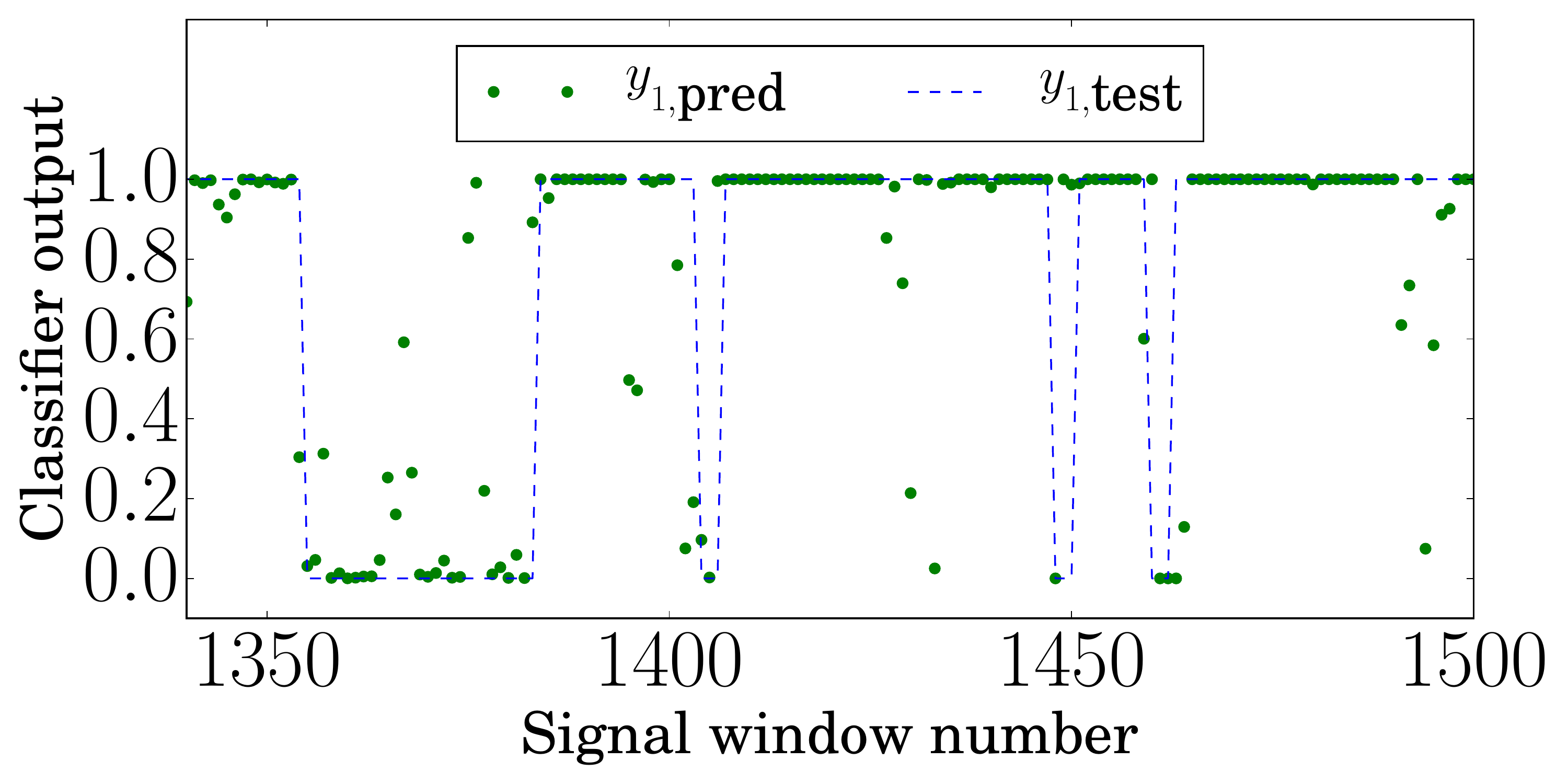}
    \caption{Wavelet}
    \label{fig:subfig:b}
  \end{subfigure}
  \caption{ROC, precision-recall, and classifier outputs over test data for \ref{fig:subfig:a}: STFT with 256 Fourier coefficients and \ref{fig:subfig:b}: wavelet with 256 scales. Target prediction for a range of signal windows is given by the blue dotted line, with actual predictions denoted by green dots. Each prediction is generated over $w_1 = 10$  samples -- a window  of 320\,ms.%
  } 
    \label{fig:results}

\end{figure}

Firstly, training the neural networks on wavelet features shows a consistent relative improvement compared to training on STFT features. We attribute the improved receiver operating characteristic curve (ROC) area to the network producing better estimates of the uncertainty of each prediction. As a result, a greater range of the detector output $0\leq y_i \leq1$ is utilised. This is best represented by the contrast in smoothness of the ROC curves, as well as the spread of predictions visible for the classifier test output in Figure \ref{fig:results}. 

Secondly, the addition of the convolutional layer provides a significant increase in every performance metric compared to the MLPs. Therefore, omitting the specific locality constraint of the CNN degrades performance.

Thirdly, the CNN trained on wavelet features is able to perform classifications  with $F_1$ score, precision-recall (PR) and ROC areas, far exceeding the results obtained with traditional classifiers. This is despite using an elaborate hand-tuned feature selection scheme that cross-validates both PCA and RFE to extract salient features. By also comparing the lower achieving CNN conditioned on STFT features, we note that both the feature representation and architecture add critical value to the detection process. 

Finally, median filtering the CNN's predictions conditioned on the wavelet features considerably boosts performance metrics, allowing our algorithm to outperform human experts. By using a median filter kernel (of 1 second) that represents the smoothness over which human labelling approximately occurred, we are able to compare performance with human expert labelling. Since human labels were supplied as absolute (either  $y_i = 1, y_i = 0$), an incorrect label incurs a large penalty on the ROC and precision-recall curve areas. This results in a far exceeding ROC area of 0.970 for the CNN-wavelet network, compared to 0.873, 0.901 and 0.874 of the three human experts respectively. However, even raw accuracies are comparable, as indicated by the near identical $F_1$ score of the best hand label attempt and our filtered algorithm. Further algorithmic improvements are readily attainable (e.g. classifier aggregation and temporal pooling), but fall beyond the scope of this paper.


\subsection{Visualising Discriminative Power}
\label{subsec:visual}
\begin{figure}[t]
\centering

    \includegraphics[page = 1, width=1.\linewidth]{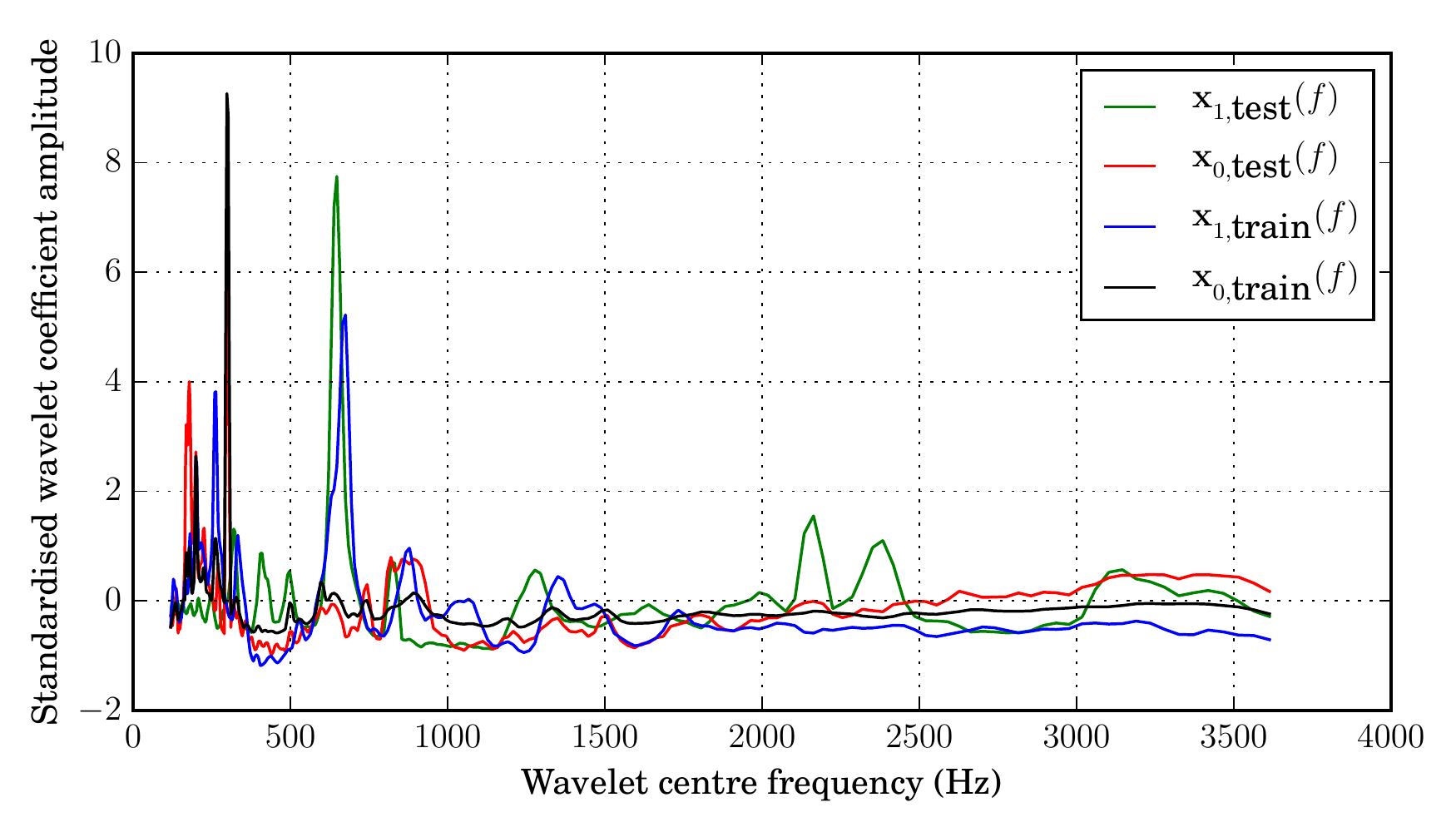}
\caption{Plot of standardised wavelet coefficient amplitude against centre frequency of each wavelet scale for the top 10\,\% predicted outputs over a test dataset. The learned spectra $\mathbf{x}_{i,\text{test}}(f)$ for the highest $N$ scores closely match the frequency characteristics of the labelled class samples $\mathbf{x}_{i, \textrm{train}}(f)$.}
\label{fig:Resultsmeans}
\end{figure}

In the absence of data labels, visualisations can be key to understanding how neural networks obtain their discriminative power. To ensure that the characteristics of the signal have been learnt successfully, we compute the frequency spectra $\mathbf{X}_i(f)$ of samples that maximally activate the network's units. We collect the highest $N$ predictions for the mosquito class, $\hat{y}_1$, and non-mosquito class, $\hat{y}_0$, respectively.
The high-scoring test data forms a tensor $\mathbf{X}_{i,\text{test}} \in \mathbb{R}^{N\times 256  \times 10}, i = \{0,1\}$, which is the concatenation of $N$ spectrogram patches. The frequency spectra are then computed by taking the ensemble average across the patches and individual columns as follows:
\begin{equation}
\mathbf{x}_{i,\text{test}}(f) = \frac{1}{10}\frac{1}{N}\sum_{j=1}^{10} \sum_{k=1}^{N} X_{ijk}, \ \mbox{where} \ X_{ijk} \in \mathbb{R}^{256}.
\end{equation}
Similarly, we compute spectra $\mathbf{x}_{i,\text{train}}(f)$ for the two classes from the $N_s$ labelled training samples.
We make our spectra zero-mean and unit-variance in order to make direct comparisons between the high-scoring test spectra for each class $\mathbf{x}_{i,\text{test}}(f)$, and their reference from the training set $\mathbf{x}_{i,\text{train}}(f)$. The resulting test spectrum for the mosquito class ($\mathbf{x}_{1}(f)$, Figure \ref{fig:Resultsmeans}) shows a distinct frequency peak around 650\,Hz. This peak clearly matches the audible frequency of the mosquito, confirming that the network is making predictions based on learnt features of the true signal. The same holds true for the noise spectra ($\mathbf{x}_{0}(f)$), which is dominated by a component around 300\,Hz. A mismatch between learnt and labelled spectra would raise warning flags to the user, suggesting the network may for example be learning to detect the noise profile of the microphones used for data collection rather than the mosquito flight tones.

\section{Conclusions}
\label{sec:Conclusion}




This paper presents a novel approach for acoustic classification of free-flying mosquitoes in a real-world, data-scarce scenario. We show that a convolutional neural network outperforms generic classifiers such as random forests and support vector machines commonly used in the field. The neural network, trained on a raw wavelet spectrogram, also outperforms traditional, hand-crafted feature extraction techniques, surpassing any combination of alternative feature-algorithm pairs. Moreover, we conclude that the addition of a convolutional layer results in performance gains over non-convolutional neural networks with both Fourier and wavelet representations. With the further addition of rolling median filtering, the approach is able to improve on human expert labelling.


Furthermore, our generic feature transform allows us to visualise the learned class representation by back-propagating predictions made by the network. We thus verify that the network correctly infers the frequency characteristics of the mosquito, rather than a peculiarity of the recording such as the microphone noise profile. Future work will generalise our model to multiple classes, such as individual mosquito species, and deploy our algorithm in a physical device to allow for large-scale collection of data. 


\subsubsection{Acknowledgements.}
This work is part-funded by a Google Impact Challenge award. Ivan Kiskin is sponsored by the the AIMS CDT (\url{aims.robots.ox.ac.uk}). This work is part of the HumBug project (\url{humbug.ac.uk}), a collaborative project between the University of Oxford and Kew Gardens.

%
%

\bibliography{ecml}

%







\end{document}